\documentclass[10pt,twocolumn,letterpaper]{article}

\usepackage{cvpr}
\usepackage{times}
\usepackage{epsfig}
\usepackage{graphicx}
\usepackage{amsmath}
\usepackage{amssymb}
\usepackage{subcaption}
\usepackage[nolist]{acronym}
\newacro{lstm}[LSTM]{Long Short-Term Memory}
\newacro{lsta}[LSTA]{Long Short-Term Attention}
\newacro{clstm}[ConvLSTM]{Convolutional Long Short-Term Memory}
\newacro{cnn}[CNN]{Convolutional Neural Network}
\newacro{dnn}[DNN]{Deep Neural Network}
\newacro{dl}[DL]{Deep Learning}
\newacro{tavlad}[TA-VLAD]{Top-Down Attention VLAD}
\newacro{vlad}[VLAD]{Vector of Locally Aggregated Descriptor}
\newacro{cam}[CAM]{Class Activation Map}
\newacro{rnn}[RNN]{Recurrent Neural Network}
\newacro{gru}[GRU]{Gated Reccurrent Unit}
\usepackage[pagebackref=true,breaklinks=true,letterpaper=true,colorlinks,bookmarks=false]{hyperref}

 \cvprfinalcopy % *** Uncomment this line for the final submission

\def\cvprPaperID{****} % *** Enter the CVPR Paper ID here

\ifcvprfinal\fi
\begin{document}

%%%%%%%%% TITLE
% \title{Deep Neural Models for Action Recognition: A Neurophysiological Analysis}
\title{An Analysis of Deep Neural Networks with Attention for Action Recognition from a Neurophysiological Perspective}

% \author{First Author\\
% Institution1\\
% Institution1 address\\
% {\tt\small firstauthor@i1.org}

% For a paper whose authors are all at the same institution,
% omit the following lines up until the closing ``}''.
% Additional authors and addresses can be added with ``\and'',
% just like the second author.
% To save space, use either the email address or home page, not both
% \and
% Second Author\\
% Institution2\\
% First line of institution2 address\\
% {\tt\small secondauthor@i2.org}
% }

\author{Swathikiran Sudhakaran$^{1,2}$ and Oswald Lanz$^{1}$\\ 
	$^{1}$Fondazione Bruno Kessler, Trento, Italy\\
	$^{2}$University of Trento, Trento, Italy\\
	{\tt\small \{sudhakaran,lanz\}@fbk.eu}
}

\maketitle
%\thispagestyle{empty}

%%%%%%%%% ABSTRACT
\begin{abstract}
   We review three recent deep learning based methods for action recognition and present a brief comparative analysis of the methods from a neurophyisiological point of view. We posit that there are some analogy between the three presented deep learning based methods and some of the existing hypotheses regarding the functioning of human brain.
\end{abstract}

%%%%%%%%% BODY TEXT
\section{Introduction}

% intro of human visual system; attention mechanism

% short intro of dnns; attention in dnn

% brief comparison of hvs and dnns

% contributions

Human visual system have the remarkable capability to accurately recognize an object present in a scene within a very short span of time, in the order of milliseconds~\cite{thorpe1996speed}. This is achieved even in the presence of wide range of identity preserving transformations such as rotation, shift in spatial position, changes in the color, size and view. Several studies have been conducted to understand the mechanism underlying this achievement and has led to several hypotheses, some of which are yet to be proved.

Computer vision researchers have tried to develop systems that can emulate the performance of human visual systems. Some of these approaches are inspired by the hypotheses and understandings developed by neuroscientists based on their study of the primate visual system. The most notable approach among these is the neocognitron~\cite{fukushima1982neocognitron} based on the primate visual model proposed by Hubel and Wiesel~\cite{hubel1977ferrier}. The neocognitron inspired the development of \acp{cnn}~\cite{lecun1998gradient} which revolutionised the area of \ac{dl} and resulted in the development of \acp{cnn} than can rival human performance in image recognition task~\cite{he2016deep, szegedy2017inception, cadieu2014deep}.

Recently, neuroscientists have started to analyze \acp{dnn} to obtain more detailed understanding of the functioning of primate visual systems by studying on the similarities of the representations generated by both systems. These studies have confirmed that the representations of the visual scene generated by \acp{cnn} are similar to the ones generated in the brain. Similar objects are found to be nearer while different objects are found to be farther in this representational space in both systems~\cite{cadieu2014deep}. Further studies have also confirmed that the ventral stream of the visual system which is responsible for object recognition has a hierarchical structure for generating visual representation of the visual scene in the form of light entering the eyes, similar to the hierarchical structure of \acp{cnn}~\cite{eickenberg2017seeing, kheradpisheh2016deep, yamins2016using}.

This extended abstract tries to continue this study from an analytical point of view by comparing existing hypotheses about the functioning of the visual system in primates to the improvements obtained by recent \ac{dl} approaches after adopting these hypotheses. The contributions  include a brief review of our recent works \cite{2018aiia,2018bmvc,2019cvpr} for action recognition from videos; an analysis of the above papers from a neurophysiological point of view; and an attempt to compare them with some of the hypotheses developed by neuroscientists regarding the functioning of the brain.

\section{Computer Vision Perspective}

	\begin{figure*}[t]
% 		\centering
		\begin{subfigure}[b]{0.12\textwidth}
			\includegraphics[width=94px, height=55px]{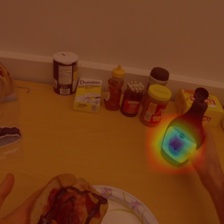}
		\end{subfigure} \hskip 13mm
		\begin{subfigure}[b]{0.12\textwidth}
			\includegraphics[width=94px, height=55px]{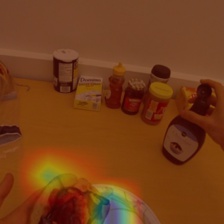}
		\end{subfigure} \hskip 13mm
		\begin{subfigure}[b]{0.12\textwidth}
			\includegraphics[width=94px, height=55px]{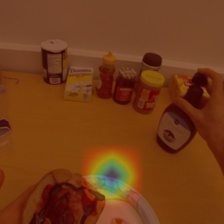}
		\end{subfigure} \hskip 13mm
		\begin{subfigure}[b]{0.12\textwidth}
			\includegraphics[width=94px, height=55px]{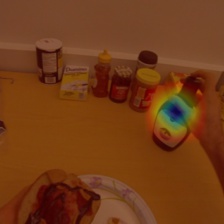}
		\end{subfigure} \hskip 13mm
		\begin{subfigure}[b]{0.12\textwidth}
			\includegraphics[width=94px, height=55px]{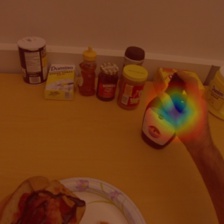}
		\end{subfigure}
		\\[1mm] %\vskip 0.01mm
		\begin{subfigure}[b]{0.12\textwidth}
			\includegraphics[width=94px, height=55px]{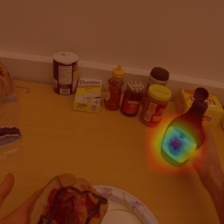}
		\end{subfigure} \hskip 13mm
		\begin{subfigure}[b]{0.12\textwidth}
			\includegraphics[width=94px, height=55px]{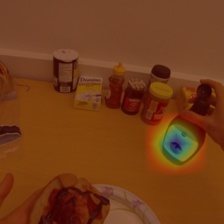}
		\end{subfigure} \hskip 13mm
		\begin{subfigure}[b]{0.12\textwidth}
			\includegraphics[width=94px, height=55px]{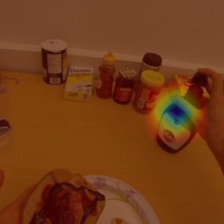}
		\end{subfigure} \hskip 13mm
		\begin{subfigure}[b]{0.12\textwidth}
			\includegraphics[width=94px, height=55px]{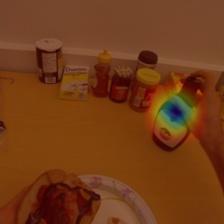}
		\end{subfigure} \hskip 13mm
		\begin{subfigure}[b]{0.12\textwidth}
			\includegraphics[width=94px, height=55px]{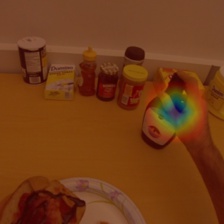}
		\end{subfigure}
		\caption{Attention maps of some frames in GTEA~61 dataset for the action class \texttt{close\_chocolate}. Top row: Ego-RNN, second row: LSTA}
		\label{fig:fig_res_ego_lsta}
	\end{figure*}

\subsection{Object-centric Attention (Ego-RNN)}

In our paper~\cite{2018bmvc}, we present a \acf{cnn}-\acf{rnn} architecture that is trained in a weak supervision setting to predict the raw video-level activity-class label associated with the clip. Our \ac{cnn} backbone is pre-trained for generic image recognition and augmented on top with an attention mechanism that uses class activation maps for spatially selective feature extraction. The memory tensor of a \ac{clstm} then tracks the discriminative frame-based features distilled from the video for activity classification. Our design choices are grounded to fine grained activity recognition because: (i) Frame-based activation maps are not bound to reflect image recognition classes, they develop their own representation classes implicitly while training the video-level classification; (ii) \ac{clstm} maintains the spatial structure of the input sequence all the way up to the final video descriptor used by the activity classification layer, thus facilitating the spatio-temporal encoding of objects and their locations into the descriptor as they develop into the activity over time.

% {\huge Sample images}

\subsection{\ac{lsta}}

In the method proposed in~\cite{2018bmvc}, the attention maps are generated independently for each frame. This can result in the network attending to different regions in adjacent frames. In order to address this limitation,
we derive \ac{lsta}, a new recurrent neural unit that augments \acs{lstm} with built-in recurrent spatial attention and a revised output gating. The first enables \ac{lsta} to attend the feature regions of interest while the second constraints it to expose a distilled view of internal memory. Our study also confirms that it is effective to improve the output gating of recurrent unit since it does not only affect prediction overall but controls the recurrence, being responsible for a smooth and focused tracking of the latent memory state across the sequence. This output pooling applies attention on the \ac{rnn} memory, thereby enabling the network to localize on the relevant spatio-temporal patterns present in the video. Fig.~\ref{fig:fig_res_ego_lsta} shows the attention map generated by Ego-RNN and \ac{lsta} on a video sequence from GTEA~61 dataset.

% {\huge sample images}

\subsection{\ac{tavlad}}

\begin{figure*}[t]
% 		\centering\
    	\begin{subfigure}[b]{0.12\textwidth}
			\includegraphics[width=94px, height=55px]{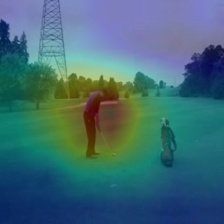}
		\end{subfigure} \hskip 13mm
		\begin{subfigure}[b]{0.12\textwidth}
			\includegraphics[width=94px, height=55px]{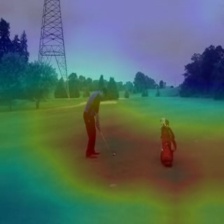}
		\end{subfigure} \hskip 13mm
		\begin{subfigure}[b]{0.12\textwidth}
			\includegraphics[width=94px, height=55px]{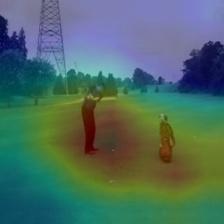}
		\end{subfigure} \hskip 13mm
		\begin{subfigure}[b]{0.12\textwidth}
			\includegraphics[width=94px, height=55px]{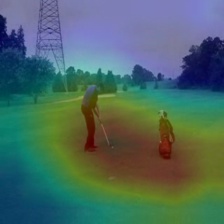}
		\end{subfigure} \hskip 13mm
		\begin{subfigure}[b]{0.12\textwidth}
			\includegraphics[width=94px, height=55px]{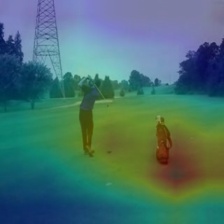}
		\end{subfigure}
		\\[1mm]
		\begin{subfigure}[b]{0.12\textwidth}
			\includegraphics[width=94px, height=55px]{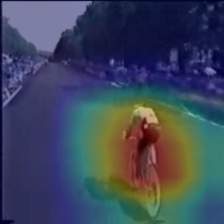}
		\end{subfigure} \hskip 13mm
		\begin{subfigure}[b]{0.12\textwidth}
			\includegraphics[width=94px, height=55px]{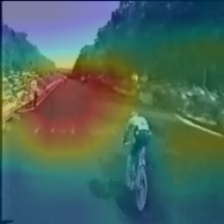}
		\end{subfigure} \hskip 13mm
		\begin{subfigure}[b]{0.12\textwidth}
			\includegraphics[width=94px, height=55px]{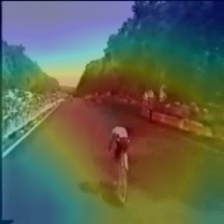}
		\end{subfigure} \hskip 13mm
		\begin{subfigure}[b]{0.12\textwidth}
			\includegraphics[width=94px, height=55px]{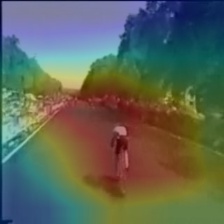}
		\end{subfigure} \hskip 13mm
		\begin{subfigure}[b]{0.12\textwidth}
			\includegraphics[width=94px, height=55px]{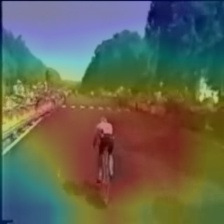}
		\end{subfigure}
		\caption{Attention maps for some frames in HMDB51 dataset. Top row: action class \texttt{golf}, second row: \texttt{ride\_bike}}
		\label{fig:fig_res_tavlad}
	\end{figure*}

Our recently published paper~\cite{2018aiia} presents an end-to-end trainable deep architecture that integrates top-down spatial attention with temporally aggregated \acs{vlad} encoding for action recognition in videos. \ac{tavlad} uses (i) class specific activation maps obtained from a deep \ac{cnn} pre-trained for image recognition as the spatial attention mechanism, a (ii) latent cluster representation of the feature space, obtained using \ac{vlad} encoding, and (iii) \acp{gru} for temporal encoding in the cluster space. \ac{tavlad} can be trained end-to-end using video-level annotations, that is, the parameters of (i) and (iii) together with the compact representation of feature space (ii) are learned from videos paired with action class labels. Fig.~\ref{fig:fig_res_tavlad} shows the attention map generated by the network on some of the frames in HMDB51 dataset.

% \ac{tavlad} is a method proposed by Sudhakaran \& Lanz~\cite{2018aiia} for action recognition from third-person videos. Their method uses the attention mechanism proposed in \cite{2018bmvc} for weighting relevant regions. However, instead of applying the weighted features obtained from a \ac{cnn} to a \ac{rnn} for temporal encoding, they build an architecture that is similar in functionality to the multiple pathways hypothesis. They use \ac{vlad} encoding of the features obtained from a \ac{cnn}. The feature descriptor corresponding to each of the \ac{vlad} cluster centers are applied to a network of \ac{gru} layers whosew parameters are shared. This is equivalent to how information flow through multiple streams in the brain.

% Even with wide range of identity preserving transformations. Two stream hypothesis states two separate visual systems are responsible for vision, ventral stream and dorsal stream.~\cite{goodale1992separate}. Ventral stream is responsible for object recognition and detection while dorsal stream is responsible for localization.

% Representation space generated by CNNs and human brain are similar; images with similar objects are closer and with different objects are far in the space ~\cite{cadieu2014deep}

\section{Neurophysiological Perspective}
\subsection{Top-down Attention}
% ego-rnn and LSTA

% Human brain is considered to have two types of attention mechanisms, namely bottom-up attention and top-down attention, that enable them to perform complex visual recognition tasks with ease~\cite{ungerleider2000mechanisms}. 

Studies on the human brain have shown that there is a limit to the number of objects that can be processed simultaneously~\cite{duncan1984selective}.
As a result, the brain selects the relevant regions in the scene to generate an effective representation. This is achieved by the attention mechanism present in the brain. Studies have confirmed that the human brain employs two types of attention mechanisms to select relevant regions present in the scene, namely bottom-up attention and top-down attention~\cite{ungerleider2000mechanisms}. Bottom-up attention is triggered by the salient features of the scene such as color and shape whereas top-down attention is based on the prior information present in the brain which results in a bias to select some regions over others.

In \cite{2018bmvc,2018aiia,2019cvpr}, we apply top-down attention on the \ac{cnn} features obtained from each frame to weight the relevant regions present in the frame. The top-down attention is generated from \acp{cam} obtained from a \ac{cnn} pre-trained for object classification. In the networks, each frame is first applied to an imagenet pre-trained \ac{cnn} to obtain a class-category score. The \ac{cam} of the class-category with the highest class score is then used to generate the attention map. This has some analogy to the top-down attention mechanism in primate brain which selects regions in the scene based on the internal bias and goals. Empirical studies have shown that weighting the regions present in the scene in this way improves the action recognition performance of the network.

Majority of the existing studies comparing the representational similarities of \acp{cnn} and primate brain consider the object recognition task. Comparative studies on the representations generated by the brain on action recognition task and \ac{cnn}-\ac{rnn} architectures like ours could shed some light on how spatio-temporal information is processed in the brain and may assist in the further development of effective action recognition techniques. Such a study could also help reveal and explain the benefit of output pooling introduced in \acf{lsta}~\cite{2019cvpr}. 
%  This could also shed some light in how spatio-temporal information is processed in the brain.

% Why this is top-down? because the attention map is generated by first passing the frame through a cnn, classify it and then the attention is generated. This way, the network, processes each frame, goes down and then generate the attention leveraging all the information about the scene that was ingrained in it during the training stage.

%  The first one is bottom-up attention which uses.... 

% The second one is top-down attention which uses prior information about the scene to focus on relevant objects present in the scene. (internal goals, bias in the network, study on face/non-face classification~\cite{tu2017network})

\subsection{Multiple Pathway Hypothesis}
% \ac{tavlad}

% However, no solid theory on how the information is represented in the brain white matter to recognize one object from another; that is, to retrieve the name or label of the object. One hypothesis presented from studies conducted on category-specific semantic deficit is that different regions are responsible for identifying different semantically similar categories(objects): classifier in CNN~\cite{mahon2009concepts}. 

Multiple pathway hypothesis states that there are several parallel information streams in the brain that carry information from one region to the other for further processing. It is assumed that these streams are weighted with different values and that there might be complex interactions between these streams which results in the final representation of the scene in the inferio-temporal (IT) cortex of the brain~\cite{warrington1987categories, nassi2009parallel}.

% There are multiple sensory channels that carry information in the brain. They have different weighting values and complex interaction between each other which results in a particular semantic representation~\cite{warrington1987categories, nassi2009parallel}. 
In \ac{tavlad}~\cite{2018aiia}, we encode the temporal evolution of the features corresponding to each of the cluster centers separately, using a network of \ac{gru} layers. This is comparable to the multiple pathway hypothesis proposed in the primate visual system. Experiments with a single \ac{gru} layer that encodes the flattened feature descriptor obtained by combining all the cluster centers significantly reduced the performance of the network. On top of this, the top-down attention allows in focusing on the relevant regions in the video, specifically the objects present in the scene. This same approach of encoding the cluster representation using multiple streams could be further investigated in the context of \ac{lsta} \cite{2019cvpr}.

\section{Conclusion}
In this extended abstract, we presented three recent works based on deep learning for addressing the problem of action recognition. We also made an analytical comparison of the proposed methods with the existing hypotheses and understandings regarding the functioning of the human visual system. From the comparative study, it is seen that the application of attention mechanism is beneficial for improving the action recognition task. However the presented works apply only top-down attention on the \ac{cnn} features while the primate brain makes use of both bottom-up and top-down attention mechanisms for focusing onto the relevant objects or regions in the scene. Recently, Tu \etal~\cite{tu2018relating} found that there is a dynamic switching between bottom-up and top-down attentions during dynamic decision making process, which shows that \acp{dnn} should also leverage both the attention mechanisms for improving their performance.

{\small
\bibliographystyle{ieee}
\bibliography{egbib}
}

\end{document}